\documentclass[letterpaper, 10 pt, conference]{ieeeconf}  % Comment this line out if you need a4paper

\IEEEoverridecommandlockouts                              % This command is only needed if 
\usepackage{amsmath,graphicx}
\usepackage{amsfonts}
\usepackage{comment}
\usepackage{hyperref}
\usepackage{float}
\usepackage{threeparttable}

\usepackage{mwe}

\title{\LARGE \bf
Cerebral Palsy Prediction with Frequency Attention Informed Graph Convolutional Networks}

\author{Haozheng Zhang$^{1}$, Hubert P. H. Shum$^{2}$ and Edmond S. L. Ho$^{3}$% <-this % stops a space
\thanks{*This work was supported in part by the Royal Society (Ref: IES\textbackslash R2\textbackslash 181024 and IES\textbackslash R1\textbackslash 191147)}% <-this % stops a space
\thanks{$^{1}$Haozheng Zhang is with Durham University, the United Kingdom
        {\tt\small haozheng.zhang@durham.ac.uk}}%
\thanks{$^{2}$Hubert P. H. Shum is with Durham University, the United Kingdom. Corresponding author.
        {\tt\small hubert.shum@durham.ac.uk}}%
\thanks{$^{3}$Edmond S. L. Ho is with Northumbria University, the United Kingdom
        {\tt\small e.ho@northumbria.ac.uk}}%
}

\begin{document}

\maketitle
\thispagestyle{empty}
\pagestyle{empty}

%%%%%%%%%%%%%%%%%%%%%%%%%%%%%%%%%%%%%%%%%%%%%%%%%%%%%%%%%%%%%%%%%%%%
\begin{abstract}
Early diagnosis and intervention are clinically considered the paramount part of treating cerebral palsy (CP), so it is essential to design an efficient and interpretable automatic prediction system for CP. We highlight a significant difference between CP infants' frequency of human movement and that of the healthy group, which improves prediction performance. However, the existing deep learning-based methods did not use the frequency information of infants' movement for CP prediction. This paper proposes a frequency attention informed graph convolutional network and validates it on two consumer-grade RGB video datasets, namely MINI-RGBD and RVI-38 datasets. Our proposed frequency attention module aids in improving both classification performance and system interpretability. In addition, we design a frequency-binning method that retains the critical frequency of the human joint position data while filtering the noise. Our prediction performance achieves state-of-the-art research on both datasets. Our work demonstrates the effectiveness of frequency information in supporting the prediction of CP non-intrusively and provides a way for supporting the early diagnosis of CP in the resource-limited regions where the clinical resources are not abundant.
\end{abstract}

%%%%%%%%%%%%%%%%%%%%%%%%%%%%%%%%%%%%%%%%%%%%%%%%%%%%%%%%%%%%%%%%%%%%%%%%%%%%%%%%
\section{INTRODUCTION}
\label{sec:intro}
General Movement Assessment (GMA)~\cite{Einspieler2005} is being widely used clinically for the early prediction of cerebral palsy (CP). However, targeted GMA training for clinicians is a time-consuming and resource-consuming task. As a result, only a small but increasing number of clinicians have received this training in the UK and Australia~\cite{Graham2019}. Furthermore, the process also requires manual inspection of the infant movement and is prone to subjective assessment. Early studies applied machine learning techniques (e.g. support vector machine, random forest) and the optical flow-based video analysis method to propose the automated GMA systems~\cite{Orlandi2018,Ihlen2020}. But these works still require manual labelling of infant joint positions.  
Some later studies focus on the analysis of frequency domain data. Stahl \textit{et al}.~\cite{Stahl2012} used an optical flow-based approach to assess infant movements and then applied wavelet frequency analysis to evaluate the time-dependent trajectory signals in optical flow data. Rahmati \textit{et al}.~\cite{Rahmati2016}
applied a motion segmentation algorithm to extract motion data from each limb in the infant video and then classified the infants' movements with features obtained by frequency analysis. %However, this method was significantly limited by the accuracy of its motion segmentation algorithm.

Recent deep learning-based systems achieved impressive performance in CP infants movement prediction. McCay \textit{et al.}~\cite{McCay2019} proposed a fully connected deep learning network and four Convolutional Neural Network (CNN)-based deep learning architectures to classify the abnormal movements of CP infants by using the histogram of joint orientation 2D and joint displacement 2D features, achieved the highest prediction accuracy of $91.67\%$ on the MINI-RGBD dataset~\cite{Hesse2018}. Zhu~\cite{Zhu2021} further applied the channel attention mechanism on the 2D-CNN model to interpret the CP prediction outcome on the same dataset. However, the robustness and generality of their proposed method have not been fully evaluated since the results are obtained from a single small dataset.

%Nguyen \textit{et al.}~\cite{Nguyen2021} developed a Spatial-Temporal Attention-based  

%However, their work did not use the frequency information of human motion highly relevant to CP and limited by the robustness of small training data size. 

%(by using the histogram of joint orientation 2D and joint displacement 2D features). We adopt these models as the baseline approaches in our research.

%The development of deep learning provides another research perspective for automatic infants CP diagnosis. McCay et al. ~\cite{McCay2019} proposed a fully connected deep learning network and four Convolutional Neural Network-based deep learning architectures to classify the abnormal movements of CP infants %(by using the histogram of joint orientation 2D and joint displacement 2D features). We adopt these models as the baseline approaches in our research.

Aiming at the significant difference in joints movement frequency between the cerebral palsy infants and the healthy group, in this article, we demonstrate a frequency-based binning mechanism and a graph convolution network to improve the performance of CP prediction with better interpretability. Firstly, we employ a pose estimation algorithm, namely Openpose~\cite{cao2019openpose} to extract the human joint position data from the R-GBD video sequences as the input to our system. Then we propose an automatic frequency-binning module suitable for videos with different frame rates to reduce data noise and the percentages of high-frequency movements information in the whole video sequence for CP prediction. The idea is inspired by both the frequency analysis-based infants CP prediction methods~\cite{Rahmati2016} and our observation. Rahmati \textit{et al}.~\cite{Rahmati2016} provided a result that comparing with very low or high-frequency ranges, the middle-to-low frequency range data showed more differences between the healthy group and the CP group. In addition, we found that the infants' joint position data in the high-frequency domain is mainly caused by data noise, such as the misdetected joint position by Openpose. 
We validate our system on the MINI-RGBD dataset~\cite{hesse2018b} and the RVI-38 dataset~\cite{McCay2022}. The MINI-RGBD dataset has been widely used for CP classification performance comparison in the previous work~\cite{McCay2019,McCay2019a,Wu:MovCompIdx,Sakkos2021}, including synthetic video sequences of 12 normal and CP infants. The RVI-38 is a recently collected dataset for a more challenging CP prediction task, with a larger size of data captured during routine clinical care. %in daily situations. 
Experimental results show that our system achieves state-of-the-art CP prediction performance on both of the dataset and allows users to interpret the weights of movement frequencies of different joints in our prediction system.

Our contributions are as follows:
\begin{itemize}
\item We interpret the Cerebral Palsy prediction in the joint movement frequency domain by the attention module. In addition, we designed a new frequency-binning module that can be applied to both deep learning and machine learning networks for videos with different frame rates to improve the CP prediction performance.

%We propose to adapt ST-GCN for the application of CP prediction. We tackled the challenge of training a deep neural network with a small dataset through a number of network structure designs.
\item  We propose a novel frequency attention informed graph convolutional network (FAIGCN) for CP prediction from consumer-grade RGB-D videos. Our system achieves state-of-the-art research on two datasets with strong robustness. 
\item We open our source code for validation and further development:  \href{https://github.com/zhz95/FAIGCN}{https://github.com/zhz95/FAIGCN}
\end{itemize}

\section{Dataset Processing}
\label{sec:format}

%\subsection{The MINI-RGBD Dataset}
We verify our models on the Moving INfants In RGB-D synthetic dataset (MINI-RGBD)~\cite{hesse2018b} and RVI-38 dataset. 
\subsection{The MINI-RGBD Dataset}
MINI-RGBD was generated by registering and rendering the synthetic Skinned Multi-Infant (SMIL) model~\cite{Hesse2018} to the RGB-D sequences of real-world moving infants recorded in the hospital. All 12 RGB-D video sequences were captured when the infants were half-year-old. The MINI-RGBD dataset is a popular open resource relating to infants CP as it consists of realistic shape, texture and movement. It also provides precise ground truth while anonymizing the data by replacing the raw video frames with computer graphics rendered frames. We further obtained the annotation of each video sequence shared by~\cite{McCay2019a}, which indicates the presence (i.e. labelled as ``normal'') or absence (i.e. labelled as ``abnormal'') of fidgety movements in the video by an independent medical expert using the GMA method~\cite{Einspieler2005}. %The dataset was labelled into two categories. The ``abnormal'' labels are used if some video frames show a lack of fidgety movements. The ``normal'' labels are for video sequences with fidgety movements demonstrated by the infants. 
%Although the size of this dataset is small, its publicity allows us to compare our model with other existing studies~\cite{McCay2019,McCay2019a,Wu:MovCompIdx,Sakkos2021}.
\subsection{The RVI-38 Dataset}
The RVI-38 dataset was collected from a part of routine clinical care at the Royal Victoria Infirmary (RVI) in Newcastle upon Tyne, UK. There are 38 RGB-D video sequences of different infants between 12-21 weeks in the RVI-38 dataset. All videos were captured by a consumer-grade handheld camera (Sony DSC-RX100 with a resolution of 1980x1080 and the 25FPS frame rate). The length of videos ranges between 40 seconds and 5 minutes, with an average length of 3 minutes and 36 seconds. The camera was set above the baby, and the infant's movement was photographed from top to bottom. All videos were annotated using the GMA method by two experienced assessors. The annotations indicate the presence (i.e. labelled as ``normal'') or absence (i.e. labelled as ``abnormal'') of fidgety movements in the video.

\subsection{Data Preprocessing}
For more effective CP predictions, we extract 2D skeleton features from the video sequences. We apply OpenPose ~\cite{cao2019openpose} for pose estimation due to its high accuracy in detecting the posture of the infants, and it is less sensitive to variations on the appearance. %(e.g. clothing, body shape, image colour). 
%As an estimation result, 
OpenPose returns the 2D coordinates $(x,y)$ for 18 human joint landmarks and a confidence score $C$ for each joint estimation. However, for joints that are self-occluded or without clear visual features, OpenPose would not be able to deduce their position, and zero values would be returned as the joint positions. As this may impact the performance of the prediction system, we propose to preprocess the data by replacing the zero values in frame $f$ with the linear interpolation of neighbouring non-zero frames.

In order to overcome the overfitting in small size dataset, we implement several processes. 1) We calculate the global normalization of joint positions frame by frame to reduces the infant's global translation. To achieve this, we set the center of the triangle of the neck and two hip joints as the global origin, then relocate each joints by the relative distance between joints. 2) To normalize the x-direction and y-direction pose features, we align the line between the global orgin and the neck joint with the y-axis and keep the neck joint above the global origin.

\begin{figure*}[!t]
\centering
\includegraphics[width=0.80\textwidth]{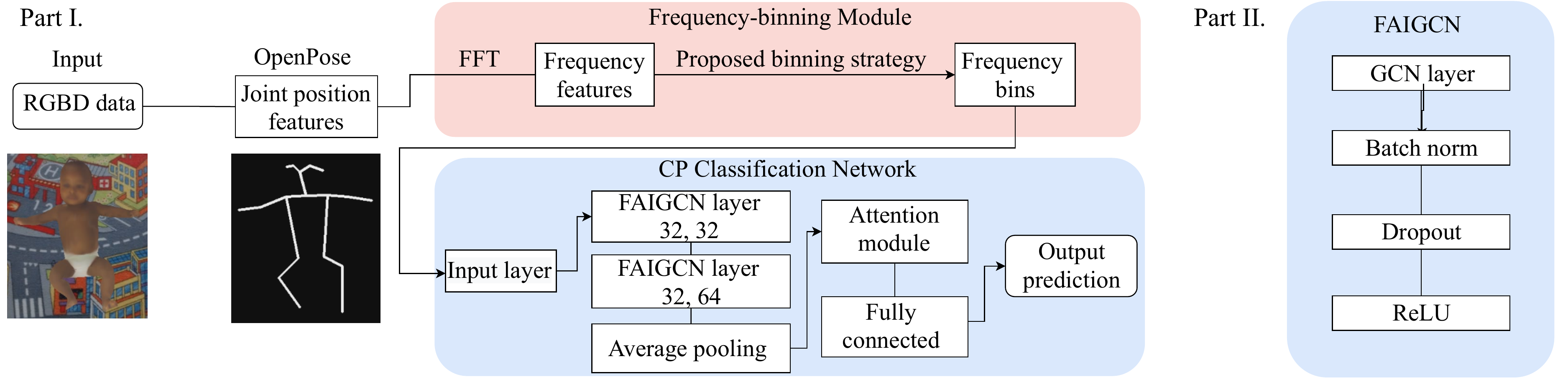}
\caption{The overview of our proposed framework. Part I is the overall network architecture, Part II is the design of each FAIGCN layer.}
\label{fig_sim}
\end{figure*}

\begin{figure}[H]
\centering
\includegraphics[width=0.2\textwidth]{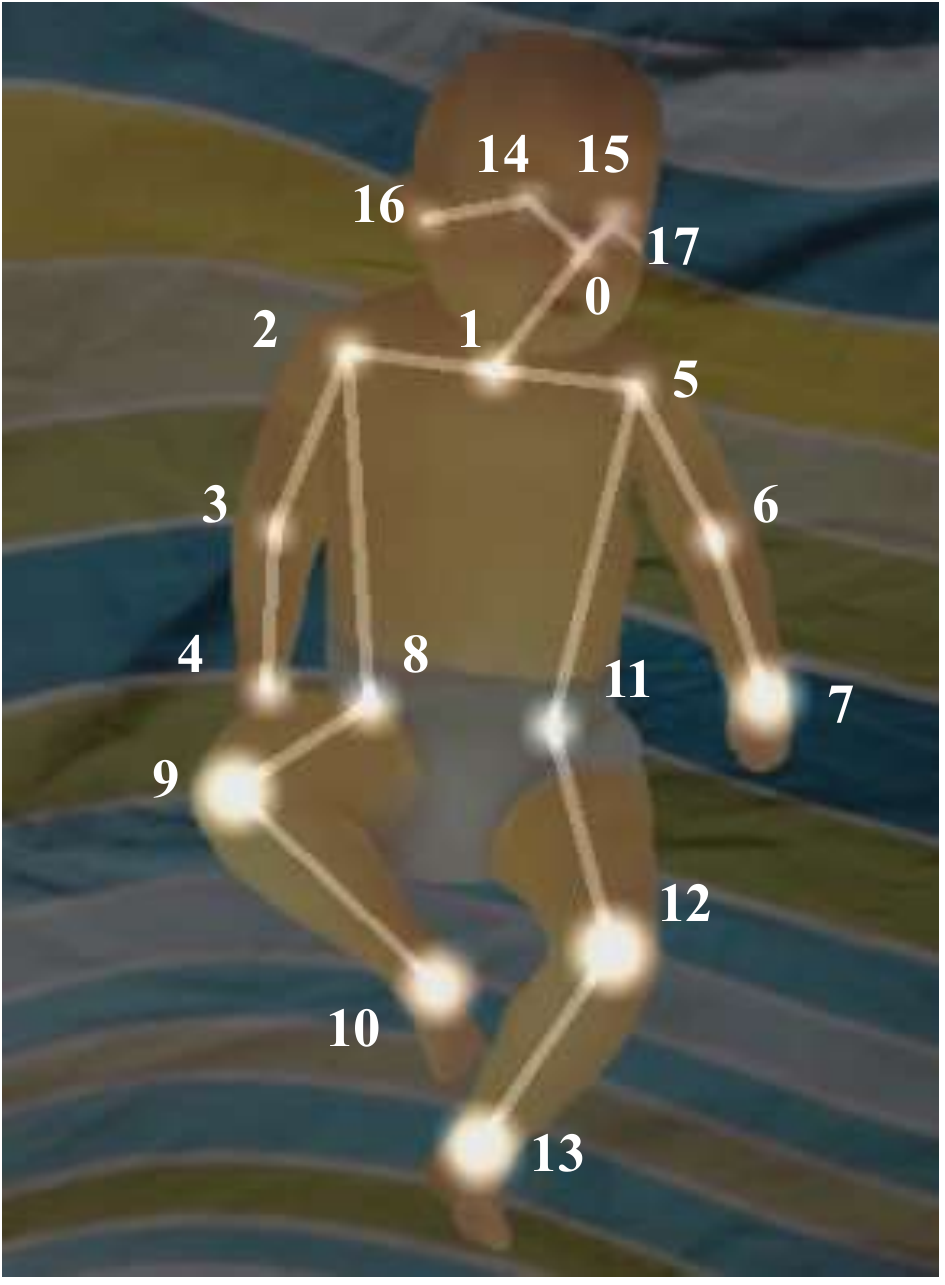}
\caption{An example frame of 18-joints Openpose~\cite{cao2019openpose} posture layout for infant kicking in the MINI-RGBD dataset ~\cite{Hesse2018}. The size of the light point represents the size of the attention value, that is, the importance of the movement frequency of the joint to our network in CP prediction at that frame.}
\label{openp}
\end{figure}

\section{Frequency Attention Informed GCN System}
The proposed system consists of two parts (seen in Fig.\ref{fig_sim}): (1) The frequency-binning module transforms the input joint movement features into the frequency domain, then filters high-frequency information to make our prediction network focus on low-to-mid infants movement frequencies. (2) The proposed Frequency Attention Informed GCN for CP prediction and interpretation. Fig. \ref{openp} shows an example of attention visualisation.
%In this section, we present our novel solution that combines the use of a frequency-binning module and a Spatial-temporal Graph Convolutional Network (ST-GCN) for CP prediction.

%In this section, we present our novel solution that combines the use of a frequency-binning module and a Spatial-temporal Graph Convolutional Network (ST-GCN) for CP prediction.

\subsection{The Frequency-binning Module}
Given the infants' joint movement sequences as input, we propose using frequency operations on joint position data for CP prediction. It is motivated by two observations. Firstly, the body movement frequencies of healthy infants are different from infants who suffered from CP~\cite{Rahmati2016}, and the low-frequency range information of body movement is more critical for fidgety movements (FMs). FMs are moderate speed movement of the neck, trunk and limbs with different accelerations in various directions~\cite{Ferrari2002}. Previous work~\cite{Ferrari2002,Einspieler2016} has shown that the absence of FMs is an essential distinguishing feature of CP infants from healthy infants. Frequency-binning can filter the high-frequency (e.g., above 6 HZ) information of joint position data after FFT, thus making the classification network focused on low-to-mid (e.g., 0-5 HZ) frequency infant movements without eliminating raw data. %\textcolor{red}{(approximate the original time series to the sum of several low and mid-frequency series, without high-frequency series)}
  %Such an observation supports that using a good frequency model would improve the data representation for better predictions. 
Secondly, the infant movements frequencies are generally low, and the high-frequency range from the joint position data is mainly due to data noise, such as the misdetected joint position and the video capture error from the datasets. %It would be beneficial to apply filter processing to minimize the impact of high-frequency noise.

As a solution, we design a frequency-binning module that retains the critical frequency of the joint position data while filtering the noise. The module employs Fast Fourier Transform (FFT) to convert the time series of joint positions into the frequency domain, then applies frequency binning to obtain the motion frequency information mainly distributed in the low-to-mid band. This module is adaptable for videos with a frame rate between 24 FPS to 60 FPS and is suitable for both DNN or machine learning-based classification models. The core of the module is the binning strategy, in which we design a formula to use finer bins for the more crucial low-to-mid frequency and coaster bins for higher frequency. 
%\ref{sec:experiment}.

%while low frequency movements refer to the slow translational motion. Due to postural and movement disorders caused by CP, abnormal infants’ movements are likely to be more concentrated in a middle-to-low frequency range than normal infants, and because a large proportion of cerebral palsy infants were reported to be accompanied by epileptic seizures, the movements of abnormal infants may be different from that of normal infants in a very high frequency range.

%More specifically, the procedure of the module is as follows: 

\subsubsection{Fast Fourier Transform (FFT)}
We apply Bluestein’s FFT algorithm \cite{Bluestein1970}, a discrete Fourier transform algorithm, on all 2D joints movements time series to transform original joints position features into the frequency domain and obtain the %corresponding 
frequency components:
%. The mathematical definition of FFT is referred to discrete Fourier transform, which is defined as:
\begin{equation}
X_k = \sum_{n=0}^{N-1} x_n e^{\frac{-i2\pi kn}{N}}, \quad k=0,\dots,N-1
\label{FFT}
\end{equation}
where $x_n$ is a time series, $e^{\frac{i2\pi}{N}}$ is a primitive $N^{th}$ root of 1.

\subsubsection{The Binning Strategy}
We propose a data binning strategy to emphasize the importance of low-frequency information of the joint position data. Under the strategy, the width of the bins are different - smaller width bins are used for low-frequency range %(e.g. $1$ to $5$ HZ$)$ 
and increasingly larger-width bins for higher frequency range:
%to include the integer units of frequency components.
%\begin{equation}
%bin-size = \{ 0.2Hz if freq<6Hz
%              0.4Hz if 6Hz<freq<8Hz
%              0.6Hz if freq>8Hz
%bin-size = \begin{cases}
%         $0.2Hz$, &\text{if $freq<6Hz$,}
%        \\
%         $0.4Hz$, &\text{if $6Hz \leq freq<8Hz$,}
%        \\
%        $0.6Hz$, &\text{if $freq \geq 8Hz$}.
%        \end{cases}
%\end{equation}
\begin{equation}
%bin-size = e^{freq} / 
b_{n} = \begin{cases}
         $Round$( b_{0} \cdot c^n ), &\text{if $b_{n} \cdot c^n< 3$,}\\
         $Ceiling$( b_{0} \cdot c^n ), &\text{if $b_{n} \cdot c^n \geq 3$,}
        \end{cases}
        \label{eqbin}
\end{equation}
where $b_n$ is the width of the $n^{th}$ bin, $b_0=1$, and $c$ is a controllable parameter. Note that the width of each bin needs to be an integer as FFT is a discrete (i.e. integer-based) system.
This equation takes the round of the bin width when the width is less than three units to increase the density of the bins in low-to-mid frequency. According to the characteristics of rapid exponential growth, this function distinguishes the density of the middle frequency band and the high-frequency band for bins with a width greater than two units by rounding up the value greater than three units. 
%Our purpose is to increase the weight of low-to-mid frequency components and control the total number of bins to avoid leading an excessive number of temporal features (maximum 300) into the CP prediction network.
Empirically, as shown in Sec \ref{sec:experiment}, we achieve the best prediction accuracy when $c = 1.00264$ for the 25 FPS videos. The parameter $c$ could be automatically generated for the best binning results by achieving the highest CP prediction performance for different datasets. %We both apply this binning strategy on the joint position features and the corresponding confidence scores $C$ to keep them in the same dimension.  %and further provide an automatic empirical algorithmic to calculate this parameter value.

%\subsubsection{Inverse Fast Fourier Transform (IFFT)}
%The mean value of each bin is adopted to be new frequency component. Then, IFFT is applied to map the features in frequency domain back to spatial domain:
%We reconstruct the joint trajectories by converting the features in the frequency domain back to the spatial domain using IFFT. This allows the more critical frequency components are being emphasized in the FFT step while enabling the analysis of joint movements' temporal information, which is crucial to the CP prediction network.
%Here, the mean value of each bin is adapted to be a new frequency component. 
%\begin{equation}
%x_n = \frac{1}{N}\sum_{k=0}^{N-1} X_k e^{i2\pi kn/N}, \quad n\in \mathbb{Z} 
%\label{iFFT}
%\end{equation}
%We applied IFFT to map frequency components back to the time domain to keep the temporal information of joint position features, which is an important part of  our CP prediction network. Conversely, frequency components cannot be directly used since they do not include any temporal information.

As a result, after being processed by the frequency-binning module, input joint position data are transformed into the frequency domain and endowed with an important characteristic: low-to-mid frequency information occupies a significantly more prominent emphasis. 

\subsection{The CP Prediction Network}
As shown in Fig.\ref{fig_sim} , we propose a Frequency Attention Informed Graph Convolutional Network (FAIGCN) for CP prediction by classifying low-to-mid frequency band infant movement frequency features with the attention mechanism. 

\subsubsection{Frequency Attention Informed Network}

Most of the previous DNN-based studies on infants CP prediction are based on traditional Convolutional Neural Networks (CNN). However, traditional discrete convolution from CNN can only maintain translational invariance on Euclidean data, which is not suitable for graph structure data such as the human skeletal graph generated from OpenPose~\cite{cao2019openpose}. 

Therefore, we employ a GCN~\cite{Kipf2017} to learn the infant joints dependencies from the pose graph. Inspired by \cite{yan2018}, we apply the pose graph which align with the human skeletal graph \begin{math} G = (V,E)\end{math} for interpreting which joint's movement frequency features are considered to be important in CP prediction task. In this graph, \begin{math} \{V = {v_{bi}|b=1,...,B; i=1,...,N}\}\end{math} denotes the frequencies of all joints, where \begin{math} v_{bi} \end{math} represents the \textit{b}-th frequency bin of \begin{math} i\end{math}-th joint. The edge set \textit{E} includes: (1) the intra-skeleton connection at each frequency band, \begin{math} \{v_{bi}v_{bj}|(i,j)\in K \} \end{math}, where \begin{math} K \end{math} is designed by the natural connections of human joints. (2) the inter-frequency edges which connect the frequency bins of a joint in the low-to-high frequency order, \begin{math} \{v_{bi}v_{(b+1)i}\} \end{math}.

The graph convolutional operation of FAIGCN is followed by~\cite{Kipf2017}, where the propagation rule between layers can be represented by Eq. \ref{GCN}. 
\begin{equation}
    \mathbf{H}^{(l+1)} = \sigma \left(\tilde{\mathbf{D}}^{-\frac{1}{2}}\tilde{A}\tilde{\mathbf{D}}^{-\frac{1}{2}}\mathbf{H}^{(l)} \mathbf{W}^{(l)} \right)
    \label{GCN}
\end{equation}
where \begin{math} \tilde{\mathbf{A}} = \mathbf{A}+\mathbf{I}_L \end{math} is known as the adjacency matrix of an undirected graph. \begin{math} \mathbf{I}_L  \end{math} is an \textit{L} dimensions identity matrix. \begin{math} \tilde{\mathbf{D}}_{ii} = \sum_j \tilde{\mathbf{A}}_{ij}  \end{math} and \begin{math} \mathbf{W}^{(l)}\end{math} is a learnable weight matrix specified to the layer. The nonlinear activation function \begin{math} \sigma(\cdot) \end{math} is set as \textit{ReLU} in our network.

We propose the following frequency attention-informed mechanism to learn the weight of frequency features. We aggregate the frequency features obtained from the frequency-binning module \begin{math}\{ \mathbf{h}_{1,i},\mathbf{h}_{2,i},\dots,\mathbf{h}_{B,i} \}\end{math} with attentions \begin{math}\alpha _{b,i}\end{math} by Eq. \ref{agg}
\begin{equation}
    \mathbf{v}_k = \sum_{b=1}^{B} \alpha_{b,i}\mathbf{h}_{b,i}
    \label{agg}
\end{equation}
in which the frequency attention weight \begin{math}\alpha _{b,i}\end{math} is defined as:
\begin{equation}
    %\alpha_{b,i} = \text{softmax}\left(\alpha^\mathsf{T} \mathbf{z}_{b,i}\right) 
    \alpha_{b,i} = \frac{\text{exp} \left(\sigma^{'}_n\left(\mathbf{w}_{\alpha}^\mathsf{T},  \mathbf{z}_{b,i}\right)\right)}{\sum_{b}\text{exp}\left(\sigma^{'}\left( \mathbf{w}_{\alpha}^\mathsf{T},\mathbf{z}_{b,i}\right)\right)}
    \label{alpha}
\end{equation}

\begin{equation}
    \mathbf{z}_{b,i} = tanh\left( \mathbf{W}_{z}\mathbf{h}_{b,i}\right)
    \label{h}
\end{equation}
\\
where $\sigma^{'}_n$ is an adjustable activation function as follows:
\begin{equation}
 \sigma^{'}_n =\ 
\begin{cases}
1 + \left( \frac{\mathbf{w}_{\alpha}}{\left \| \mathbf{w}_{\alpha}  \right\|	}\right)^{\mathsf{T}}  \left( \frac{\mathbf{z}_{\alpha}}{\left \| \mathbf{z}_{\alpha}  \right\|	}\right)  & ,n = 1\\
\mathbf{w}_{\alpha}^\mathsf{T}\mathbf{z}_{\alpha} & ,n = 2
\end{cases}
\end{equation}
\\
where $\mathbf{w}_{\alpha}$ and $\mathbf{W}_{z}$ are learnable parameters.
%The input feature map of proposed network is \textit{f_{in}^c}, where \textit{c} represent the dimension of the input channel. The output of each channel by convolutional operation 

\subsubsection{Network Adaptation}

%Because MINI-RGBD is a small size dataset, we have modified the network architecture for the following aspects and improved the performance of original ST-GCN in verifying small s. 

As can be seen from Fig.\ref{fig_sim}, the input layer transforms the tensor format of input frequency data to fit in the network. Then, we use two FAIGCN layers with 32, 64 output channels respectively. Each FAIGCN, in turn, consists of a GCN layer, a batch normalization layer, a dropout and a ReLU layer. The kernel sizes of FAIGCN layers $K=3,3$, and $stride = 1, 2$, respectively. We put a global pooling layer after two FAIGCN layers. We applied the average pooling as it provides the highest robustness. At the last, we put a fully connected layer to classify features for CP prediction. The optimizer is chosen as \textit{Adams}, and we train the model with \textit{batch size} $=1$, \textit{learning rate} $=0.0001$ with $0.1$ decay every 100 epoches, \textit{Max Epoch} $=500$ on the MINI-RGBD dataset;  \textit{batch size} $=4$, \textit{learning rate} $=0.001$ with $0.1$ decay every 100 epoches, \textit{Max Epoch} $= 500$ on the RVI-38 dataset.

\section{Experiments}
\label{sec:experiment}
Our experiments were run on a PC with % is developed on 
Ubuntu 18.04 and an NVIDIA GeForce RTX 3080. The total model training time on MINI-RGBD with 12 sequences is about an hour, including estimation of the joints position from RGB videos. But it only takes about 50s for the CP prediction of 1000 frames ($\sim33$s) video sequence, which can be employed in interactive-time diagnosis.

\subsection{Experimental Settings}
In this paper, we conduct the leave-one-out cross-validation among two datasets to evaluate our proposed system. This setting utilises all data and ensures that the prediction system is evaluated against unseen data. Our evaluation metrics are introduced in Sec \ref{sec:metrics}. We report the best result for each method to be consistent with several related works in  literature~\cite{Stahl2012,Einspieler2016,Rahmati2016,Sakkos2021,McCay2022}.

\subsection{Comparing with State-of-the-art Methods}
In order to evaluate the effectiveness of our system, we compare FAIGCN with the following methods:
\begin{itemize}
\item  \textbf{FCNet}~\cite{McCay2019}: This method uses fully connected deep network architectures to the Histogram of Joint Displacement 2D (HOJD2D) and Histogram of Joint Orientation 2D (HOJO2D) calculated from human joint positions. For the HOJD2D feature, the displacements of each joint are extracted every five frames and segmented into 16 bins. The feature of HOJO2D represents the joint orientation in 2D space, and the feature is also segmented into 16 bins. 
\item \textbf{Conv1D-1}, \textbf{Conv1D-2}~\cite{McCay2019}: They are two 1D convolutional neural networks, each of them consists of two 1D convolutional layers with differences in the output channel sizes. They are proposed to classify the abnormal infant movements by feature HOJO2D or HOJD2D (HOJO/D2D).
\item  \textbf{Conv2D-1}, \textbf{Conv2D-2}~\cite{McCay2019}: They are two 2D convolutional neural networks, each of them consists of two 2D convolutional layers with differences in the output channel sizes. 
\item \textbf{CANet}~\cite{Zhu2021}: This method proposes a 2D convolutional neural network with the squeeze-and-excitation channel attention module. The whole system is proposed for the CP classification task on the MINI-RGBD dataset.
\item \textbf{ST-GCN} (Spatial Temporal Graph Convolutional Network)~\cite{yan2018}: This is a graph convolutional neural network for human skeleton data (e.g. joint position).
\item \textbf{STAM}~\cite{Nguyen2021}: This is a spatial-temporal graph convolutional neural network with the attention mechanism.
\item \textbf{Ens-1}~\cite{McCay2019a}: This method uses an ensemble classifier on the fused feature of HOJO2D + HOJD2D (HOJO+D2D) with eight bins.
\item \textbf{Ens-2}~\cite{McCay2022}: This method extends \cite{McCay2019a} by fusing four pose-related features and three velocity-related features.
\item \textbf{Ens-3}~\cite{McCay:BHI2021}: This method extends \cite{McCay2019a} by extracting features at limb-level from small video segments to locate abnormal movements spatiotemporally.
\item \textbf{MCI}~\cite{Wu:MovCompIdx}: This method uses a threshold model to classify the infant CP via Movement Complexity Index (MCI), where MCI is computed by extracting the infant's limb angle features.
\end{itemize}

\subsection{Evaluation Metrics}
\label{sec:metrics}
We evaluate our system and other state-of-the-art methods by following five metrics: the prediction accuracy (AC) shows the percentage of correctly predicted individuals in the dataset; the sensitivity (SE) shows the percentage of correctly predicted positive individuals among the total number of positive individuals in the dataset; the specificity (SP) shows the percentage of correctly predicted negative individuals among the total number of negative individuals in the dataset; F1-Score evaluates the binary classification performance by calculating the harmonic mean of the precision and recall; Matthews Correlation Coefficient (MCC)~\cite{Matthew1975} provides a reliable performance metric for imbalanced dataset~\cite{David2020}.
\begin{table}[H]
\caption{The comparison with state-of-the-arts on the MINI-RGBD}
\label{table1}
\renewcommand\tabcolsep{5.0pt}
\centering
\scriptsize
\begin{threeparttable} 
\begin{tabular}{cccccccc}
\hline
Method&
Feature&
AC & 
SE &
SP &
F1 &
MCC
%PR &
%F1
\\
\hline
FCNet~\cite{McCay2019}& 
HOJD2D &
91.67&
\textbf{100.00}&
87.50 &
88.89 &
83.67
%1.000 &
%0.857

\\
FCNet~\cite{McCay2019}& 
HOJO2D &
83.33 &
75.00 &
87.50 &
75.00 &
62.50
%1.000 &
%0.857
\\
Conv1D~\cite{McCay2019}&
HOJO/D2D &
83.33 &
75.00 &
87.50 &
75.00 &
62.50
%0.750 &
%0.750 
\\
Conv2D~\cite{McCay2019}& 
HOJO/D2D &
83.33 &
75.00 &
87.50 &
75.00 &
62.50
%1.000 &
%0.857 
\\
Conv2D~\cite{McCay2019}& 
HOJO+D2D &
91.67&
\textbf{100.00}&
87.50 &
88.89 &
83.67
%1.000 &
%0.857 
\\
Conv2D~\cite{McCay2019}& 
Pose &
83.33 &
75.00 &
87.50 &
75.00 &
62.50
\\
CANet~\cite{Zhu2021} & 
Pose &
91.67&
\textbf{100.00}&
87.50 &
88.89 &
83.67
%1.000 &
%0.857 
\\
ST-GCN\cite{yan2018} & 
Pose &
91.67&
\textbf{100.00}&
87.50 &
88.89 &
83.67
\\
STAM~\cite{Nguyen2021} & 
Pose &
91.67&
\textbf{100.00}&
87.50 &
88.89 &
83.67
\\
\hline
Ens-1~\cite{McCay2019a} &
HOJO+D2D & 
91.67&
\textbf{100.00}&
87.50 &
88.89 &
83.67
%1.000 &
%0.857  
\\
Ens-2~\cite{McCay2022} &
Velocity &
91.67&
\textbf{100.00}&
87.50 &
88.89 &
83.67
%1.000 &
%0.857  
\\
Ens-2~\cite{McCay2022} &
Pose\tnote{*}& 
\textbf{100.00}&
\textbf{100.00} &
\textbf{100.00} &
\textbf{100.00} &
\textbf{100.00}
\\
Ens-2~\cite{McCay2022} &
Vel.+Pose\tnote{*} & 
\textbf{100.00}&
\textbf{100.00} &
\textbf{100.00} &
\textbf{100.00} &
\textbf{100.00}
%1.000 &
%0.857  

%1.000 &
%0.857 
\\
Ens-3~\cite{McCay:BHI2021} &
HOJO+D2D & 
\textbf{100.00}&
\textbf{100.00} &
\textbf{100.00} &
\textbf{100.00} &
\textbf{100.00}
%1.000 &
%0.857  

%1.000 &
%0.857  
\\
\hline
MCI~\cite{Wu:MovCompIdx} & 
Limb angle &
91.67&
\textbf{100.00} &
87.50 &
88.989 &
83.67
%0.857 &
%0.889
\\
\hline
\hline
FAIGCN &
Motion Freq. &
\textbf{100.00}&
\textbf{100.00} &
\textbf{100.00} &
\textbf{100.00} &
\textbf{100.00}
\\
\hline
\end{tabular}

\begin{tablenotes}
\footnotesize
\item[*] The Pose and velocity features here fuses several hand-crafted 
\\features including HOJOD2D 
 %\item[**] my website is ... 
\end{tablenotes}
\label{tab1}
\end{threeparttable}
\end{table}
\subsection{Result Comparison}
We report the prediction results on MINI-RGBD and RVI-38 datasets on Table \ref{tab1} and Table \ref{tab2} respectively. From our evaluation, we propose the following observations:
(1) Our FAIGCN system outperforms the state-of-the-art DNN based methods in both datasets. Comparing with other non-DNN based methods, our system also achieves state-of-the-art performance in two datasets. (2) We can observe the advantage of the attention mechanism in DNN as CANet outperforms Conv2D-Pose from Table \ref{tab1} and Table \ref{tab2}, and CANet outperforms Conv1D/Conv2D, STAM outperforms ST-GCN from Table \ref{tab2}. (3) From Table \ref{tab2}, it can be seen that ST-GCN, STAM and FAIGCN outperform all CNN-based methods (i.e. Conv1D, Conv2D and CANet), which confirms the advantage of using graph structure to analyse human pose data. (4) We notice that the methods that use early fusion on features outperform those using only a single kind of feature. It can be seen by comparing Conv2D-HOJO/D2D with Conv2D-HOJO+D2D, or comparing Ens-2-Velocity/Pose with Ens-2-Velocity+Pose in both tables. Therefore, we consider fusing our movement frequency features with other features in future work. (5) An interesting finding is that the Machine learning-based methods such as Ens-1, Ens-2 and Ens-3 outperform DNN-based methods except for FAIGCN. On the one hand, it shows the superiority of hand-craft features in the classification tasks; On the other hand, it inspires us to compare FAIGCN with machine learning-based methods with the same features, seen in the Sec. \ref{abla} below.

\begin{table}[H]
\renewcommand\tabcolsep{5pt}
\caption{The comparison with state-of-the-arts on the RVI-38}
\label{table2}
\centering
\scriptsize
\begin{threeparttable}
\begin{tabular}{cccccccc}
\hline
Method&
Feature&
AC & 
SE &
SP &
F1 &
MCC
%PR &
%F1
%1.000 &
%0.857 
\\
\hline
Conv2D~\cite{McCay2019}& 
Pose &
81.58 &
33.33 &
90.63 &
36.36 &
25.85
\\
CANet~\cite{Zhu2021} & 
Pose &
86.84 &
66.67 &
90.63 &
61.54 &
53.89
\\
ST-GCN~\cite{yan2018} & 
Pose &
89.47 &
66.67 &
93.75 &
62.50 &
60.42
\\
STAM~\cite{Nguyen2021} & 
Pose &
92.11 &
\textbf{83.33} &
93.75 &
76.92 &
72.51
\\
\hline
Ens-1~\cite{McCay2019a} &
HOJO+D2D & 
94.74 &
\textbf{83.33} &
96.88 &
83.33 &
80.21

\\
Ens-2~\cite{McCay2022} &
Velocity &
94.74 &
\textbf{83.33} &
96.88 &
83.33 &
80.21
\\
Ens-2~\cite{McCay2022} &
Pose\tnote{*} & 
94.74 &
\textbf{83.33} &
96.88 &
83.33 &
80.21
\\
Ens-2~\cite{McCay2022} &
Vel.+Pose\tnote{*} & 
\textbf{97.37} &
\textbf{83.33 }&
\textbf{100.00} &
\textbf{90.91} &
\textbf{89.89}
\\
\hline
\hline
FAIGCN &
Motion Freq. &
\textbf{97.37} &
\textbf{83.33 }&
\textbf{100.00} &
\textbf{90.91} &
\textbf{89.89}
\\
\hline
\end{tabular}

\begin{tablenotes}
\footnotesize
\item[*] The Pose and velocity features here fuses several hand-crafted 
\\features including HOJOD2D 
 %\item[**] my website is ... 
\end{tablenotes}
\end{threeparttable}
\label{tab2}
\end{table}

\subsection{Ablation study}
\label{abla}
The ablation study is divided into two parts: 

In the first part, we evaluate whether there is any adverse effect on prediction performance caused by the frequency-binning module (B.) or the attention module (A.). The corresponding result is displayed in Table \ref{tab3}. Due to the limitation of the MINI-RGND dataset's size, the contribution of the attention mechanism is not reflected significantly. But from the larger RVI-38 dataset, it can be seen that there is a significant improvement by applying the attention module or frequency binning module.

In the second part, we implement four machine learning-based methods with a proposed frequency-binning module to predict the CP using the movement frequency features from both datasets. The methods are Support Vector Machine (SVM), Decision Tree (Tree), Logistic Regression (LR) and Linear Discriminant Analysis (LDA). The ensemble of classification models Ens-1~\cite{McCay2019a}, Ens-2~\cite{McCay2022}  and Ens-3~\cite{McCay:BHI2021} are not included since the types of the ensemble classifier in Matlab was used which consists of a wide range of classifiers and handles the late-fusion internally. %ensembles are not detailed in their work. 
Besides, we validate the effectiveness and robustness of the frequency-binning module by eliminating it from each method. The results are reported in Table \ref{tab4}. We observe the advantage of using the proposed frequency-binning module as each method outperforms its variant of no frequency-binning module, except in the case of SVM in the MINI-RGBD dataset. In addition, we note that our system outperforms the implemented machine learning-based methods. It demonstrates the effectiveness of graph convolutional neural network in dealing with the same features.

\begin{table}[H]
\caption{The performance of FAIGCN and its simplified variants}
\centering
\scriptsize
\setlength{\tabcolsep}{3mm}{
\begin{tabular}{lcccccc}
\hline
\multicolumn{6}{c}{The MINI-RGBD dataset}
\\
\hline
\multicolumn{1}{c}{Method} &
AC &
SE &
SP &
F1 &
MCC
\\
\hline
FAIGCN-full &
\textbf{100.00} & 
\textbf{100.00} & 
\textbf{100.00} & 
\textbf{100.00} & 
\textbf{100.00}
\\
\hline
    w/o A. &
\textbf{100.00} & 
\textbf{100.00} & 
\textbf{100.00} & 
\textbf{100.00} & 
\textbf{100.00}
\\
    w/o B. &
91.67&
\textbf{100.00} &
87.50 &
88.89 &
83.67
\\
    w/o A. B. &
91.67&
\textbf{100.00} &
87.50 &
88.89 &
83.67
\\
\hline
\\
\\
\hline
\multicolumn{6}{c}{The RVI-38 dataset}
\\
\hline
\multicolumn{1}{c}{Method} &
AC &
SE &
SP &
F1 &
MCC
\\
\hline
FAIGCN-full &
\textbf{97.37} &
\textbf{83.33} &
\textbf{100.00} &
\textbf{90.91} &
\textbf{89.89}
\\
\hline
    w/o A. &
92.11 &
\textbf{83.33} &
93.75 &
76.92 &
72.51
\\
    w/o B. &
89.47 &
66.67 &
93.75 &
62.50 &
60.42
\\
    w/o A. B. &
86.84 &
66.67 &
90.63 &
61.54 &
53.89
\\
\hline
\end{tabular}}
\label{tab3}
\end{table}

\begin{table}[H]
\caption{The comparison with machine learning based methods and their variant without frequency-binning module}
\centering
\scriptsize
\begin{tabular}{ccccccc}
\hline
\multicolumn{6}{c}{The MINI-RGBD dataset}
\\
\hline
Methods& 
AC &
SE &
SP &
F1 &
MCC
\\
\hline

SVM   &
66.77 &
75.00 & 
62.50 &
66.67 &
35.36
\\
SVM w/o B. &
66.77 &
75.00 & 
62.50 &
66.67 &
35.36
%\textbf{1.000} &
%\textbf{1.000} 
\\
Tree    &
75.00 &
75.00 &
75.00 &
66.67 &
47.81
\\
Tree w/o B.&
66.77 &
75.00 & 
62.50 &
66.67 &
35.36
\\
LDA &
83.33 &
75.00 &
87.50 &
75.00 &
62.50
\\
LDA w/o B.&
75.00 &
75.00 &
75.00 &
66.67 &
47.81
\\
LR &
91.67&
100.00 &
87.50 &
88.89 &
83.67
\\
LR w/o B.&
75.00 &
75.00 &
75.00 &
66.67 &
47.81
\\
\hline
FAIGCN-full &
\textbf{100.00} &
\textbf{100.00} & 
\textbf{100.00} & 
\textbf{100.00} & 
\textbf{100.00}
\\
FAIGCN w/o B. &
91.67&
\textbf{100.00} & 
87.50 &
88.89 &
83.67
%1.000 &
%0.857 
\\
\hline
\\
\\
\hline
\multicolumn{6}{c}{The RVI-38 dataset}
\\
\hline
Methods& 
AC &
SE &
SP &
F1 &
MCC
\\
\hline
SVM  &
63.16 &
66.67 & 
62.50 &
66.67 &
35.36
\\
SVM w/o B.  &
55.26 &
50.00 & 
56.25 &
26.09 &
4.58
%\textbf{1.000} &
%\textbf{1.000} 
\\
Tree  &
81.58 &
66.67 &
84.38 &
53.33 &
43.78
\\
Tree w/o B. &
68.42 &
50.00 & 
71.89 &
33.33 &
17.16
\\
LDA    &
83.33 &
75.00 &
87.50 &
75.00 &
62.50
\\
LDA w/o B. &
65.79 &
66.67 &
65.63 &
38.10 &
24.09
\\
LR &
78.95 &
66.67 &
81.25 &
50.00 &
39.68
\\
LR w/o B. &
57.89 &
66.67 &
56.25 &
33.33 &
16.74
\\
\hline
FAIGCN-full &
\textbf{97.37} &
\textbf{83.33} &
\textbf{100.00} &
\textbf{90.91} &
\textbf{89.89}
%1.000 &
%0.857 
\\
FAIGCN w/o B. &
89.47 &
66.67 &
93.75 &
62.50 &
60.42
%1.000 &
%0.857 
\\
\hline
\end{tabular}
\label{tab4}
\end{table}

\begin{figure}[!t]
\centering
\includegraphics[width=0.47\textwidth]{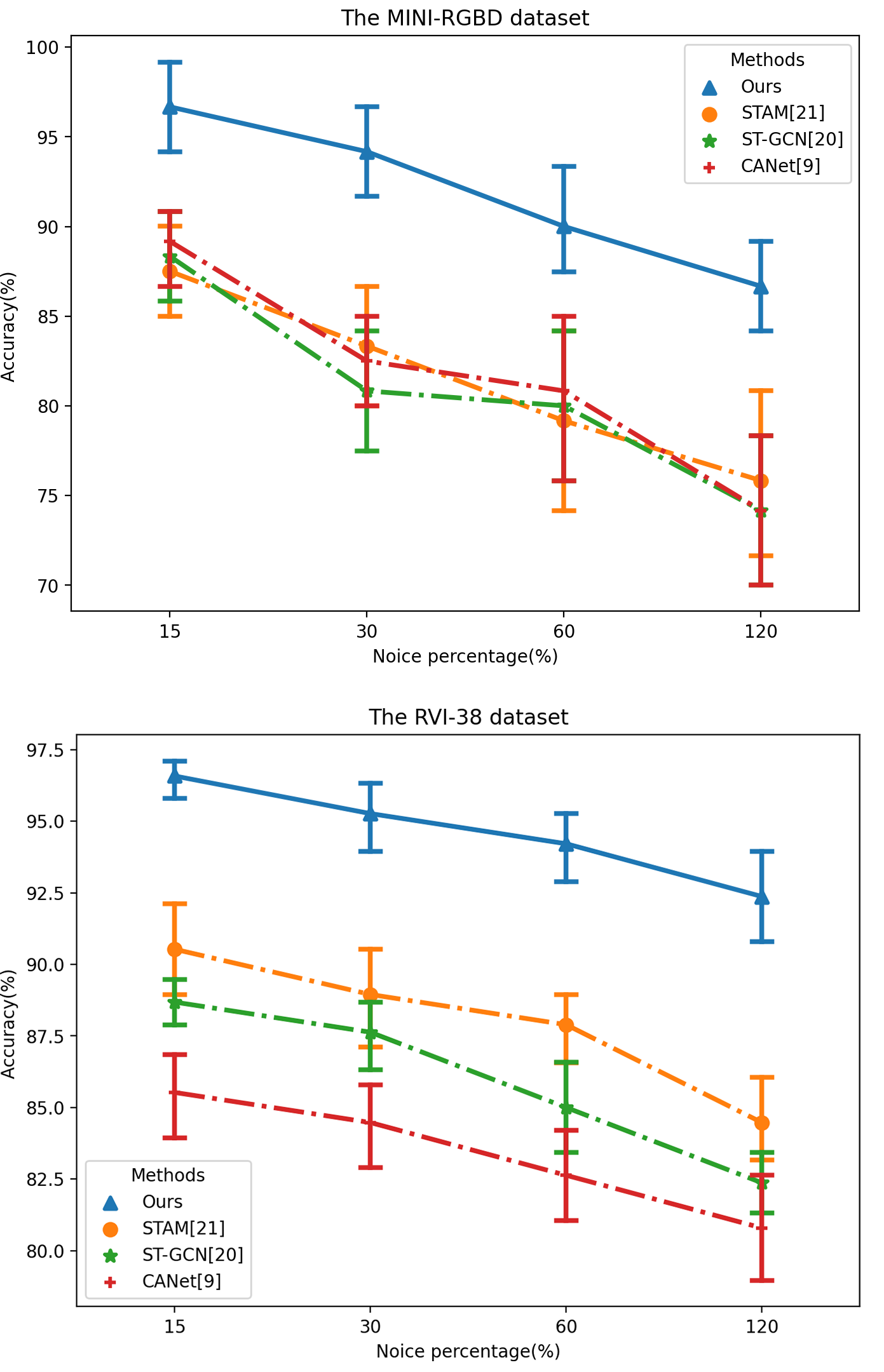}
\caption{The robustness test compared with the state-of-the-art DNN methods. The short vertical bar of each method in different noise-level denotes the accuracy range between the first quartile and third quartile among all cross-validations. The line between each bar is linked by the mean accuracy value. }
\label{exp3}
\end{figure}

\subsection{Robustness Test}
In order to evaluate the robustness of our system and other state-of-the-art DNN-based methods~\cite{Zhu2021,yan2018,Nguyen2021}, we simulate different datasets by adding different levels of Gaussian noise to the infant joint pose data. The noise level is divided into four levels from 15\% standard deviation to 120\% standard deviation of each infant's joint pose data. The tests results are displayed in Fig. \ref{exp3}. The accuracy in the y-axis is the average accuracy among ten leave-one-out cross-validations with ten different training seeds.

From Fig. \ref{exp3}, we observe that as the noise level increases, each method decreases more slowly on the RVI-38 dataset compared to the MINI-RGBD dataset, reflecting the stronger robustness brought by training the model on a larger dataset. In addition, we are seeing that the accuracy of our system shows a slower decreasing trend under different noise levels, which represents the stronger stability and robustness of our system.

\subsection{Qualitative Analysis}
\begin{figure}[!t]
\centering
\includegraphics[width=0.46\textwidth]{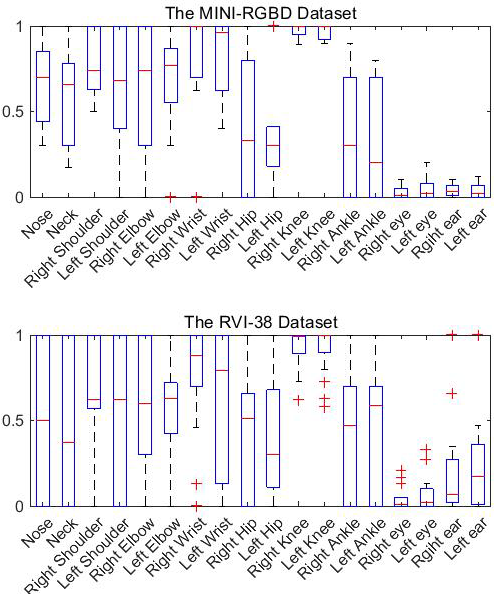}
\caption{The visualization of attention weights of different joints among all cross-validations on each dataset. }
\label{fig4}
\end{figure}

Fig. \ref{fig4} visualises the interpretability of proposed attention module by presenting the attention value of each joint among all leave-one-out cross-validations on each dataset. We observe that the attention value of `Right Knee',`Left Knee',`Right Wrist' and `Left Wrist' is significantly higher than other joints on both datasets. It indicates our system pays more attention to the movement frequencies of infants' knees and wrists, which is convincing since the movements of those joints have the most significant frequency change in the video recordings. In addition, the frequency range of `Right Eye', `Left Eye', `Right Ear' and `Left Ear' is lower than other joints significantly. One possible reason is that the self-occlusion (e.g. infant turns head) brings the noise to Openpose estimation of these joints, so that the attention module of our system lowers the weights to filter the noisy data. Besides, we notice that the attention weight range of most of the joints in the RVI-38 dataset is larger than those in the MINI-RGBD dataset. It could be caused by more information from the larger dataset.

%We generated a parameter analysis to determine the best performance value of the parameter $c$ in our custom binning equation Eq \ref{eqbin}. From the results in Table \ref{tab3}, the peak classification performance can be achieved by having around 300 bins, and therefore we designed our framework based on the empirical results. It can also be seen that having a fewer number of bins result in lower performance due to reducing overmuch variations of the movements.

%The maximum number of bins is 300, which is limited by our CP network design. Because features after the first 300 indexes will not be involved in  model training, although it still can provide an accuracy indicator. From Table 3, we can see our prediction model achieves $100\%$ accuracy when $c$ in $[1.0028,1.0029]$. However, the accuracy become lower than $91.67\%$ when number of bins less than 272, this should be caused by reducing overmuch variations of the movements by binning method. 

\section{Conclusion}
\label{sec:conclusion}

In this work, we propose a novel interpretable frequency attention informed graph convolutional network to predict cerebral palsy infants. We design a binning module for CP data to increase the weight of the low-to-mid frequency data to improve the CP prediction performance, which is adaptable for the videos with a frame rate between 24 FPS to 60 FPS and suitable for both DNN or machine learning-based classification model. Furthermore, we propose a frequency attention module to further improve the prediction performance and visualise the important joints that the network considers in CP prediction. Experimental results show the importance of low-to-mid frequency data and the effectiveness and robustness of our system in supporting the prediction of CP non-intrusively, and provides a way for supporting the early diagnosis of CP in the resource-limited regions where the clinical resources are not abundant. Future work is scheduled to apply our methods to a more extensive clinical dataset. In addition, an interesting future work could be using the transformer to interpret the frequency features better.

\section{Compliance with Ethical Standards}
The collection of the RVI-38 dataset has been ethically approved by the host organisation (Ref: 9865), the Research Ethics Committee (REC), the Health Research Authority (HRA), and Health and Care Research Wales (HCRW) (Ref: 19/LO/0606, IRAS project ID: 252317). The MINI-RGBD dataset used in this study is made open access by Fraunhofer IOSB~\cite{Hesse2018}, which had ethical approval. 

%\section{Acknowledgement}
%No funds were received for this research. The author has no relevant financial or non-financial interests to disclose.
%the Royal Society (Ref: IES$ \backslash$R2$ \backslash$181024 and IES$ \backslash  $R1$ \backslash  $191147). 

%%%%%%%%%%%%%%%%%%%%%%%%%%%%%%%%%%%%%%%%%%%%%%%%%%%%%%%%%%%%%%%%%%%%%%%%%%%%%%%%

{%\small

}
\end{document}